\documentclass[sigconf]{acmart}
\usepackage{graphicx}
\usepackage{amsmath}
\usepackage{amsthm}
\usepackage{amsfonts}
\usepackage{bm}
\usepackage{caption}
\usepackage{multirow}
\usepackage[english]{babel}
\usepackage{adjustbox}
\usepackage{subfigure}
\usepackage{booktabs}
\usepackage{array}
\usepackage{algorithm}
\usepackage{algorithmic}
\theoremstyle{definition}


\AtBeginDocument{%
  \providecommand\BibTeX{{%
    \normalfont B\kern-0.5em{\scshape i\kern-0.25em b}\kern-0.8em\TeX}}}

\copyrightyear{2021}
\acmYear{2021}
\setcopyright{acmlicensed}\acmConference[SIGIR '21]{Proceedings of the 44th International ACM SIGIR Conference on Research and Development in Information Retrieval}{July 11--15, 2021}{Virtual Event, Canada}
\acmBooktitle{Proceedings of the 44th International ACM SIGIR Conference on Research and Development in Information Retrieval (SIGIR '21), July 11--15, 2021, Virtual Event, Canada}
\acmPrice{15.00}
\acmDOI{10.1145/3404835.3463015}
\acmISBN{978-1-4503-8037-9/21/07}

\begin{document}
\fancyhead{}
\title{Explicit Semantic Cross Feature Learning via Pre-trained Graph Neural Networks for CTR Prediction}

\author{
  Feng Li$^{*}$, Bencheng Yan$^{*}$, Qingqing Long, Pengjie Wang$^{\ddagger}$, Wei Lin, Jian Xu and Bo Zheng$^{\dagger}$
 }
  \affiliation{%
  \institution{Alibaba Group}
  % \country{China}
}
 \email{{adam.lf,bencheng.ybc,lantu.lqq,pengjie.wpj,xiyu.xj,bozheng}@alibaba-inc.com, lwsaviola@163.com}

 \thanks{$*$ These authors contributed equally to this work and are co-first authors.}
 \thanks{${\ddagger}$ This author gave a lot of guidance in this work.}
 \thanks{$\dagger$ Corresponding author}
  
%   \authornote{Both authors contributed equally to this research.}
% \author{G.K.M. Tobin}`'

% \affiliation{%

% \affiliation{%

% \affiliation{%

% \affiliation{%

% \affiliation{%

% \affiliation{%

% \affiliation{%

\begin{abstract}
% Click-through rate (CTR) prediction, which aims to predict the probability of a user clicking on an ad or an item, is critical to many online applications such as online advertising and recommender systems. 
% Mining feature interactions, which has always been a core task waiting to be solved, still poses great challenges. Variety, dynamism, and huge parameters are the three core bottlenecks to be explored. 
% In this paper, we propose pre-trained Cross Feature learning Graph Neural Networks (PCF-GNN), a GNN based pre-trained model aiming at generating feature interactions in an explicit fashion and at the vector-wise level. In this work, we mark a first step in the direction of allowing pre-trained GNN models to learn an efficient and effective both explicit and implicit relationships between users and items, with few parameters, in recommendation systems. We propose a novel training loss to capture implicit feature interactions. Extensive experiments are carried out on both online billion dataset, and two public datasets, where CrossGNN shows competence in both performance and parameter-efficiency in various tasks. Our codes are available at https://drive.google.com/.

Cross features play an important role in click-through rate (CTR) prediction.
Most of the existing methods adopt a DNN-based model to capture the cross features in an implicit manner.
These implicit methods may lead to a sub-optimized performance due to the limitation in explicit semantic modeling.
 % caused by the black process of DNN.
% Although introducing the explicit semantic information of cross features
% although the explicit cross feature can 
Although traditional statistical explicit semantic cross features can address the problem in these implicit methods, such features still suffer from some challenges, including \emph{lack of generalization} and \emph{expensive memory cost}.
Few works focus on tackling these challenges.
% learning the explicit semantic information of cross features even if such semantic information can be a good signal to address the problem in the above implicit method.  
In this paper, we take the first step in learning the explicit semantic cross features and propose Pre-trained Cross Feature learning Graph Neural Networks (PCF-GNN), a GNN based pre-trained model aiming at generating cross features in an explicit fashion.
Extensive experiments are conducted on both public and industrial datasets, where PCF-GNN shows competence in both performance and memory-efficiency in various tasks.
\end{abstract}

\maketitle

% DNN explicit
% method
% two way
\section{Introduction}
\label{sec:Introduction}
Modeling the complex relationships among features is the key to click-through rate (CTR) prediction.
A major way to characterize such relationships is to introduce the cross features. 
% \footnote{we use "cross features","combinatorial features" and "feature interactions" interchangeably, as they are all used in the related literatures \cite{autoint, deepcrosssing,fm}.} 
% into CTR.
% For example, a user who is a basketball player (or a programmer) likes to click the recommended item "Nike-Air Jordan" (or digital products).
% % While a user who is a programmer may be more likely to click digital products.
For example, a user who is a basketball player  likes to click the recommended item "Nike-Air Jordan".
While a user who is a programmer may be more likely to click digital products.
These suggest that the interactions (i.e., cross features) \textit{<user\_occupation, item>} can be taken as a strong signal for CTR prediction.
In practice, such cross features have achieved remarkable results in modeling feature interactions 
% generating personalized recommendations, 
and improving the CTR performance \cite{zhou2018deepdin, wang2018billion}.

% General speaking, the way to model the cross features can be divided into two categories, i.e., implicit modeling and explicit modeling.
% (1) \textbf{Implicit Modeling.} It usually takes the embedding vectors of the involved features into a deep neural network (DNN) to model the interactions among these features \cite{qu2016product,deepcross,cheng2016wide,guo2017deepfm}.
% Since the cross features modeled by DNN refer to an implicit semantic information, we can take it as an implicit modeling.
% For example, PNN \cite{qu2016product} designs a product layer for embedding vectors of among features.
% DCN \cite{deepcross} models the high-order cross features by feeding the embedding vectors of corresponding features into a deep cross layer.
% (2) \textbf{Explicit Modeling.} A typical way to model cross feature explicitly is to count how many item clicks or calculate the click rates among the cross features from history \cite{deepcrosssing}.
% For example, we can count how many times the user who is a basketball player clicks the "Nike-Air Jordan" in the history, and take the dense counting value as one kind of the cross feature for the samples having the same user occupation and items.
% Obviously, such counting value describe the relationship among features explicitly.

There are two ways to model the cross features, i.e., implicit semantic modeling and explicit semantic modeling.
Most of the existing works focus on implicit semantic modeling.
The main idea of these works is to adopt a deep neural network (DNN) and hope the capacity of DNN can characterize the cross features in an implicit manner
% try to model the cross features by 
% taking the embedding vectors of the involved features (e.g., user occupation and item ) into 
% a deep neural network (DNN) in an implicit manner 
(e.g., Wide \& Deep \cite{cheng2016wide}, DeepFM \cite{guo2017deepfm}, DCN\cite{deepcross}, Cross-GCN \cite{feng2020cross}, and Fi-GNN \cite{fi_gnn}).
Although this kind of methods has achieved great success in cross features modeling, it is restricted to the implicit semantic modeling, 
% In other words, the learned cross features by a DNN can only capture implicit semantic information, 
and we can never guarantee the learned cross features are what we want, leading to a sub-optimized performance
% \footnote{
(Note in these methods, although the involved two features may be explicitly fed into a same DNN-based layer, the learned cross features can only contain implicit semantic information.).
% }.
Therefore explicit semantic information is an important signal to make up the limitation of the above methods.
% Meanwhile, precious few works are carried out to model the explicit semantic information of cross feature.
% In practice,

However, few works focus on learning the explicit semantic information of cross features effectively.
% Considering that explicit semantic information can be good way to  make up the limitation of the above implicit semantic modeling methods, 
% a traditional trick is to add a statistical explicit cross features (SECF) manually to 
% In practice, a simple statistical explicit cross features (SECF), are manually modeled
On the contrary, almost all the existing methods adopt simple statistical explicit semantic cross features (SESCF) manually to describe the explicit semantic information \cite{deepcrosssing}.
% Therefore, in practice, additional cross features, called statistical explicit cross features (SECF), are manually modeled in an explicit way to make up the limitation of the above implicit methods.
% \footnote{https://github.com/aister2020/KDDCUP\_2020\_Debiasing\_1st\_Place, https://github.com/loggerhead/KDD\_2012\_Track1} \cite{deepcrosssing,shan2016deep}
% (similar conclusion can be found in Section \ref{sec:Performance on CTR task (Q2)}).
Specifically, SESCF can be modeled by counting how many clicks or calculating the click rate among features from history \cite{deepcrosssing}.
For example, we can count the times that a user who is a basketball player clicks "Nike-Air Jordan" in the history, and take this counting value as the explicit semantic cross feature \textit{<user\_occupation=basketball player, item=Nike-Air Jordan>} for the samples having the same user occupation and items.
Obviously, such a counting value describes the feature relationship explicitly.
A higher value refers to a higher correlation among these features.
However, simply adopting such SESCF poses great challenges.

% \begin{figure}[t]
% % \vspace{-1.5em}
% \centering
%         \includegraphics[width = .45\textwidth]{imgs/cross_feature_modeling_example.png}
%         \caption{An example of the cross features modeling. 
%         % (a) Each row represents an impression with multiple features and one label which means click or not. (b) In the feature interaction graph, nodes represent id features, and edge weights indicate the counting values of cross features between nodes of edge such as ctr. 
%         }
%         \label{fig:cross_feature_example}
%           \vspace{-2em}
% \end{figure}

% \begin{itemize}
%     \item
 \noindent\textbf{(1) Lack of generalization}. 
% Most items recommended to users, are not bought or clicked by the users in history. Thus, there is a large requirement for predicting new pairs of <user\_feature, item\_feature>, which not appeared in the training set \cite{zhou2018deepdin}. A generalized online recommender method still remains a significant challenge.
% It is a common case where the items recommended to users, are rarely shown to the users in history.
SESCF mainly relies on the statistical counting from history and has no ability to infer the new pairs of cross features which are never shown before.
For example, the latest "Nike-Air Jordan" may be heavily browsed by the user who is a basketball player recently, and rarely shown to a user who is a programmer.
In this case, the SESCF about \textit{<user\_occupation, item>} cannot give any suggestions for the recommendation of the latest "Nike-Air Jordan" to a programmer user.
Hence, there is a large requirement for predicting new pairs and a generalized model on explicit semantic cross features still remains a significant challenge.
     % \item 

\noindent\textbf{(2) Expensive memory cost}. 
% For SECF, 
In real applications, we usually need to keep a \textit{<cross feature, counting value>} (e.g., \textit{<<basketball player, "Nike-Air Jordan">, 100>}) mapping table for online severing.
% For example, <<basketball player, "Nike-Air Jordan">, 100> may be an element for the mapping table about <user\_occupation, item\_id>.
The challenge is that maintaining such a mapping table may need excessively vast storage resources especially for a web-scale application \cite{cheng2016wide,zhou2018deepdin}.
For instance, if there are $N_1$ occupations and $N_2$ items, the number of rows in the mapping table is related to $N_1N_2$.
In other words, the corresponding memory cost is $O(N_1N_2)$ (i.e., Cartesian product).
 % where $N_1$ and $N_2$ refer to the vocabulary size of the involved features respectively.
Therefore, modeling the interaction among features which have a large-scale vocabulary (e.g., User ID) may need huge memory cost which definitely hurts the severing efficiency.

None of the existing works try to address the above challenges in the view of the explicit semantic modeling.
% To best of our knowledge, we take the first step in learning the explicit semantic of the cross features.
% to generalize the explicit cross features, 
% To tackle the generalization challenges, inspired by the great success of pre-trained models in NLP tasks \cite{devlin2018bert, brown2020language}, we design a pre-trained framework 
To tackle these challenges, we propose a Pre-trained Cross Feature learning Graph Neural Networks (PCF-GNN), a GNN \cite{long2020graph} based pre-trained model.
% aiming at 
% modeling the explicit cross features in a learning style rather than a 
% learning cross features in an explicit manner.
% Generally speaking, the 
Specifically,   
% to represent the cross features explicitly, 
we first transform the interactions among features in history to an informative graph.
Because abundant interactions can be naturally modeled as edges in a graph, and the graph can further provide us the convenience to understand the complex interactions.
In this graph, each node refers to a single feature (e.g., user\_occupation), each edge refers to a feature interaction, and the edge attributes are assigned with the counting value (i.e., SESCF).
Then a self-supervised edge-attributes prediction task is designed to pre-train a GNN (a most powerful architecture in node relationship modeling \cite{hu2019strategies, long2019hierarchical}) to allow the model to have the domain knowledge about the explicit semantic information of cross features and have the ability to infer the attributes of new edges (addressing challenge one).
Furthermore, such a design makes us free from storing the huge mapping table since we can simply infer rather than store the explicit semantic cross features (i.e., edge attributes) 
% by the corresponding features (i.e., nodes) 
(addressing challenge two).
Detailed analysis can be found in Section \ref{sec:Theoretical Analysis}.

We summarize our contributions as follows:
% \begin{itemize}
% \item 
(1) We propose PCF-GNN, a pre-trained GNN model to learn explicit semantic cross features. 
To the best of our knowledge, we take the first step in learning the explicit semantic cross features, and the pre-trained GNN is also first introduced in this area.
% \item 
(2) A pre-training pipeline is designed and a weighted loss based on interactions between features is proposed, to capture explicit semantic cross features.
% \item a edge importance and Multi-view based
% \item
(3) We carry out extensive experiments on both public and industrial datasets, where PCF-GNN shows competence in both performance and memory-efficiency.
% \end{itemize}

% \section{PRELIMINARY}
% In this section, we introduce the preliminary of our paper, including graph construction and problem definition.

% \subsection{Problem Definition}
% We  formally define the problem of click-through rate (CTR) prediction as follows:
% \begin{defn}[CTR Prediction] Let $x\in R^{n}$ denotes the concatenated features of user $u$ and item $u$, where categorical features are represented with one-hot encoding, and $n$ is the dimension of concatenated features. The problem of click-through rate prediction aims to predict the probability of user $u$ clicking on item $v$ according to the feature vector $x$.
% \end{defn}

\vspace{-0.8em}
\section{Model: PCF-GNN}
In this section, we introduce our model. 
% Fig. \ref{fig:cross_gnn} gives an overview of PCF-GNN. 
Generally speaking, we adopt a two-stage pipeline, i.e., pre-training stage (Section 2.1--2.3) and downstream task application stage (Section \ref{sec:PCF-GNN Application}).

\vspace{-0.8em}
\subsection{Graph Construction}
\label{sec:Graph Construction}
\begin{figure}[t]
\vspace{-0.8em}
\centering
        \includegraphics[width=0.7\linewidth]{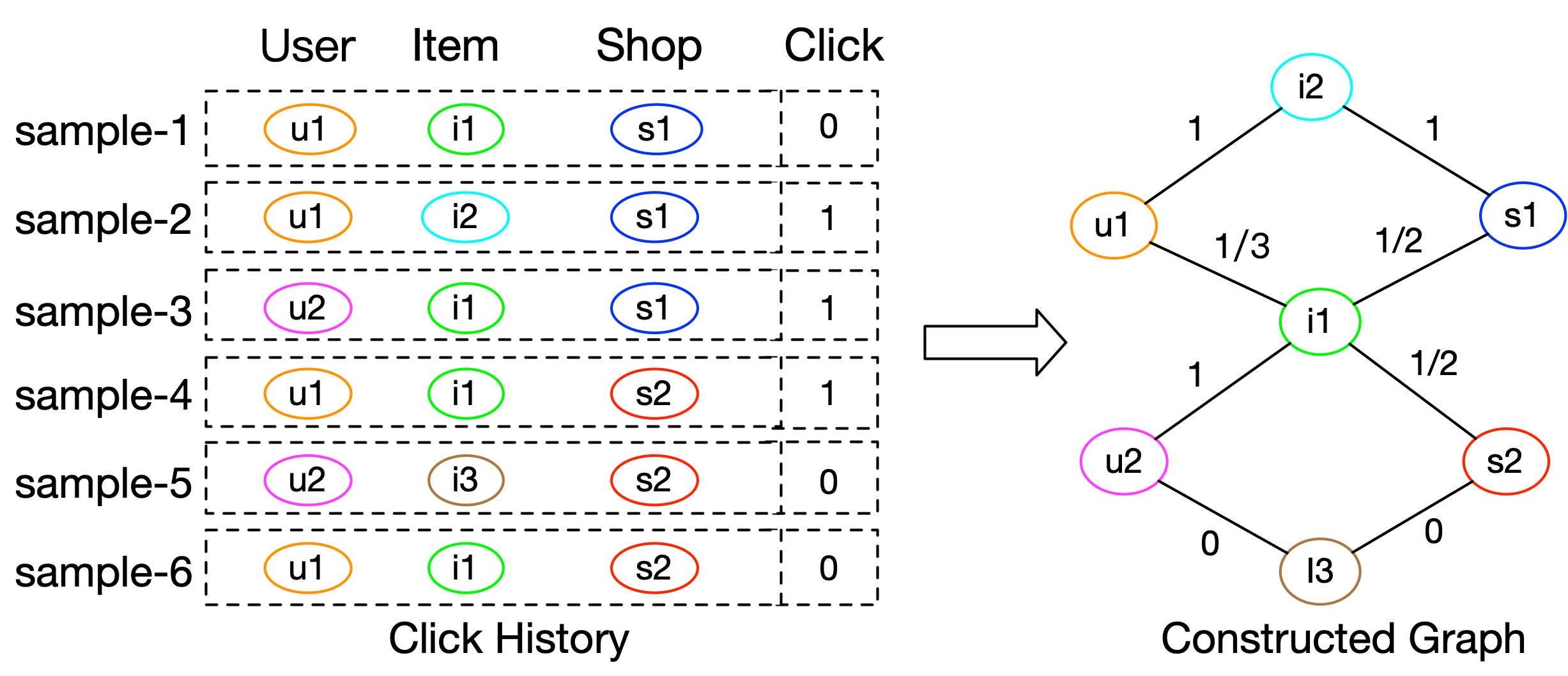}
        \vspace{-0.8em}
        \caption{An example graph constructed from history. 
        % (a) Each row represents an impression with multiple features and one label which means click or not. (b) In the feature interaction graph, nodes represent id features, and edge weights indicate the counting values of cross features between nodes of edge such as ctr. 
        }
        \label{fig:example}
          \vspace{-1em}
\end{figure}

We first construct the graph based on the interactions among features in history.
 % and then conduct pre-trained GNNs on the graphs.
As shown in Fig. \ref{fig:example}, given samples of users' click history, we take the interactions \textit{<User, Item>} and \textit{<Item, Shop>} as an example (in practice, abundant kinds of interactions are considered), and construct an informative graph where nodes refer to features, edges refer to feature interactions and the edge attributes refer to statistical explicit semantic cross features. 
% Formally,  we define a edge weight $e_{u, i}$ based on explicit relationships of user and items.
Formally, in this paper, the attribute $a_{u,i}$ of edge $e_{u,i}$ can be calculated as,
% \vspace{-0.5em}
\begin{align}
a_{u, i}& = {Count(u,i) | (click=1)}/{Count(u,i)} 
% \nonumber \\
% o_{u,i} &= C(u,i)|(click \in \{0,1\}) 
\label{eqn:graph_build}
\end{align}
% \vspace{-0.8em}
% \begin{equation}
% a_{u, i} = \frac{C(u,i) | (click=1)}{ C(u,i)|(click \in \{0,1\})}
% \label{eqn:graph_build}
% \end{equation}
where $Count(u,i)$ denotes the number of co-occurrence that features $u$ and $i$ are presented in the same history samples. 
Note $a_{u,i}$ characterizes the interactions between features $u$ and $i$ explicitly.
% In order to wipe out the outliners and noise, we use $\delta$ to filter the edges. 

\subsection{Pre-training Task Design}
\label{sec:pre-training Task Design}
% As introduced in Section \ref{sec:Introduction}, the goal of the pre-train task is to capture the cross feature explicitly.
% by a large-scale graph which are constructed from abundant history records.
% To achieve this, 
% Unlike existing methods which can only model cross feature implicitly, 
The key idea is that we need to give the domain knowledge about the explicit semantic cross features to the pre-training task. 
Specifically, since the edge attributes in the constructed graph refer to the explicit semantic cross features, we can design a self-supervised task, i.e., predicting the edge attributes.
Formally, the pre-training task can be represented as 
% \begin{equation}
$||p_{u,v} - a_{u,v}||^2$
% \end{equation}
 where $p_{u,v}$ is the output of the PCF-GNN.
 % and $a_{u,v}$ is the attribute of the edge $e_{u,v}$.
In this way, with the guidance of the edge attribute, the learned $p_{u,v}$ can explicitly characterize the cross features.

% Then the following question is (1) What is the instance of the pre-train GNN? and (2) How to design the pre-train GNN?

% \subsection{pre-train Instance Design}
% As introduce above, the pre-train task is focusing on the edge attribute prediction.
% A essential problem is how to represent each edge in the graph.
% A direct idea is to 
% General speaking, there are two ways to represent the edges, i.e., edge-level and node-level.
% The former 

\subsection{Architecture Design}
\label{sec:Architecture Design}
\begin{figure}[t]
\centering

        \includegraphics[width=0.75\linewidth]{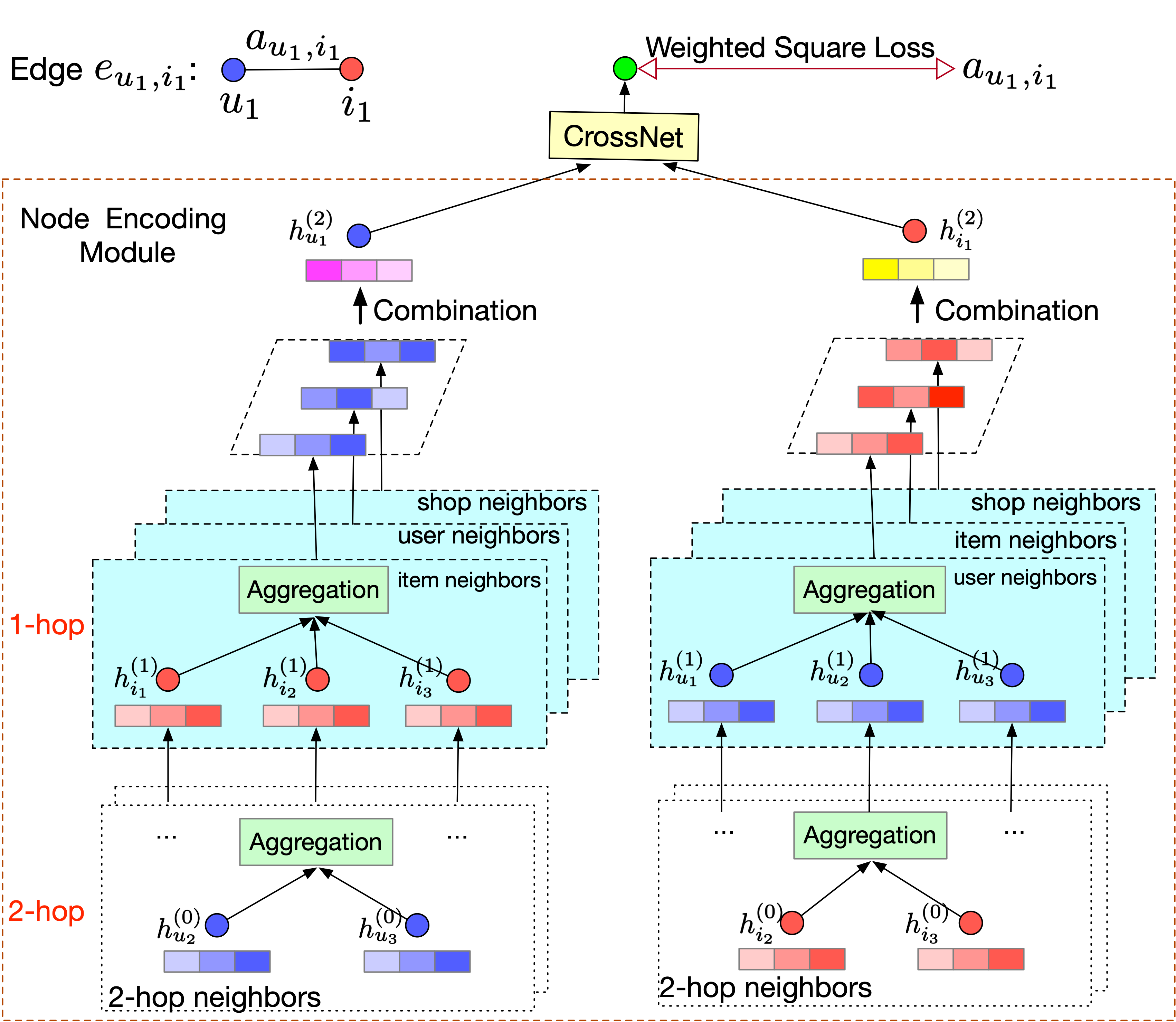}
        % \vspace{-1em}
        \vspace{-0.8em}
        \caption{An overview of the proposed PCF-GNN. 
        % Each feature performs a multi-layered neighbor query on the feature interaction graph and perform neighbor aggregation according to edge weights between current node and the neighbor nodes. Finally, the pre-trained embeddings of two features perform a muli-layered fully connect with sigmoid, then compare with edge weight using square loss. 
        }
        \label{fig:cross_gnn}
          \vspace{-2em}
\end{figure}

% As introduced above, the pre-train task is focusing on the edge attribute prediction.
% To achieve this goal, we design a GNN based 
As shown in Fig \ref{fig:cross_gnn}, the architecture of PCF-GNN can be divided into two parts, i.e., node encoding module and CrossNet.
% The former is designed to capture the node representation by neighborhoods.
% The latter transforms the node-level embedding into a edge (i.e., cross feature) space and gives a prediction for edge attribute.

\subsubsection{Node Encoding Module}
\label{sec:Node Encoding Module}
This module is a GNN which can capture the neighbor structure and learn the node embeddings.
Specifically, considering the constructed graph is a heterogeneous graph, we propose a multi-relation based aggregation function,
\begin{align}
    m_{i,r}^{(k)} &=\mathrm{AGGREGATE}\left(\left\{h_j^{(k-1)},j\in N_r(i)\right\}\right) \\
    h_{i}^{(k)} &= \mathrm{COMBINATION}\left(h_{i}^{(k-1)},\left\{m_{i,r}^{(k)}, r \in [1,R]\right\}\right),
\label{eqn:aggregate}
\end{align}
% \vspace{-1em}
where $h_i^{(k)}$ denotes the output vector for node $i$ at the $k$-th layer, 
$N_r(i)$ refers to the neighbors have the relationship $r$ with the node $i$. 
$m_{i,r}^{(k)}$ refers to the embedding of $N_r(i)$. 
The $h_i^{(0)}$ are the attributes of nodes.
For simplicity, in practice, the node attributes are defined as a learnable vector.
The $h_{i}^{(K)}$ refers to the output of node $i$ at the last layer of the node encoding module.
Actually, we take $h_{i}^{(K)}$ as the pre-trained embedding of node $i$.
% \textbf{Edge Importance and Multi-view based Aggregation}
% Furthermore, we try to take CTR informative relationships into account and propose to utilize the CTR relationships to guide the aggregation of neighbors. We utilize the scheme of importance sampling for nodes based on the edge weight $e_{u,i}$, such that the neighborhood can be aggregated in a way that illustrates \textit{homophily}.
% Formally, the $\mathrm{AGGREGATE}$ in Equ \ref{eqn:aggregate} can be rewritten as 
% \begin{align}
%     m_{i,r}^{(k)} =\mathrm{AGGREGATE}\left(\left\{\frac{e_{i,j}}{\sum_j  e_{i,j}}h_j^{(k-1)},v_j\in N_r(v_i)\right\}\right),
% \label{eqn:improved_aggregate}
% \end{align}
In this paper, We take identical $\mathrm{AGGREGATE}$ and $\mathrm{COMBINATION}$ as GraphSAGE \cite{hamilton2017inductive} while being flexible to more sophisticated methods.

\begin{figure}[t]
\centering

        \includegraphics[width=0.7\linewidth]{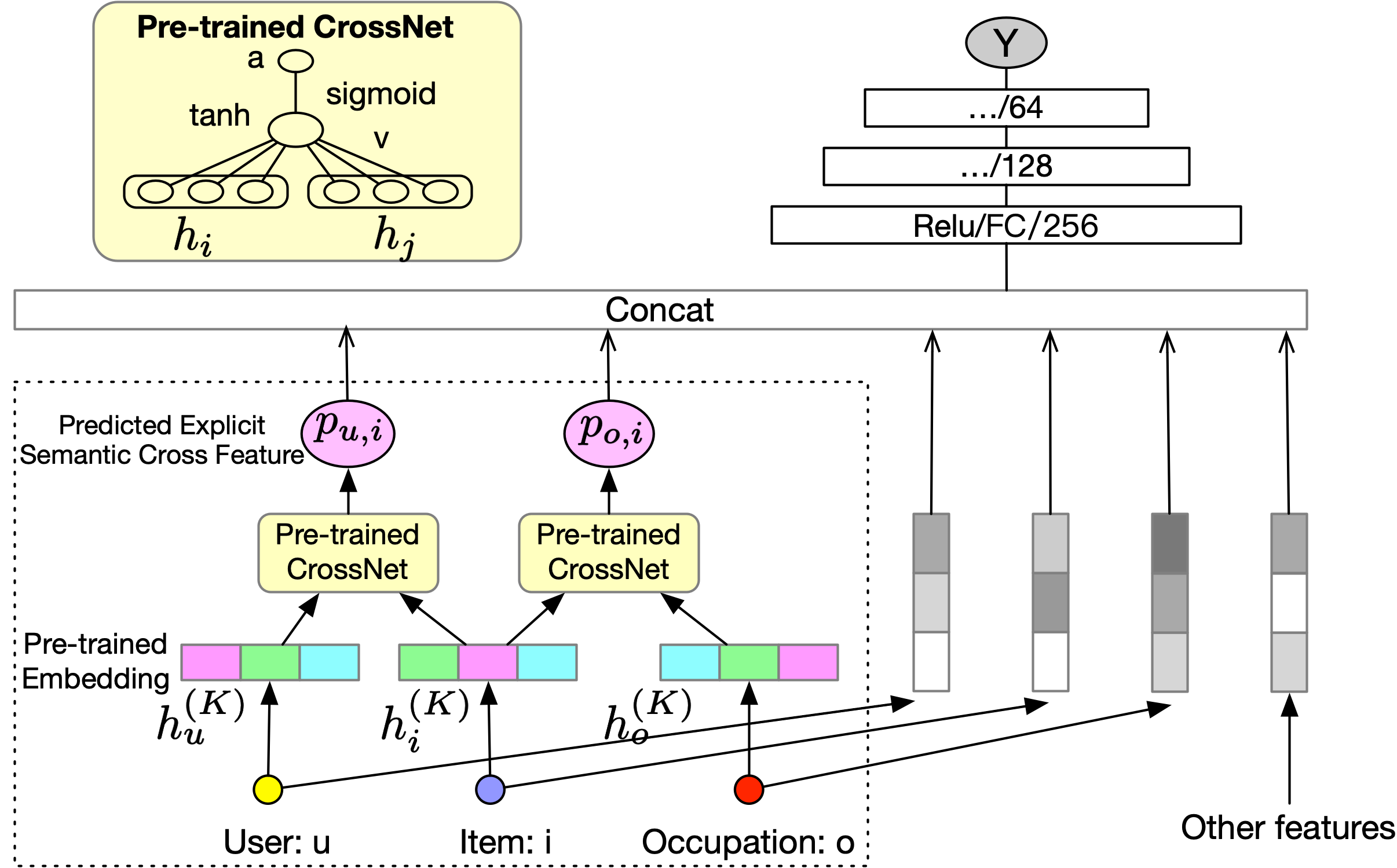}
         % \vspace{-1em}
         \vspace{-0.5em}
        \caption{PCF-GNN application on the CTR prediction task. 
        % Each id feature gets pre-trained embedding and compute counting of cross feature using pre-trained CrossNet. Finally, the counting of cross feature and other features are concatenated as inputs for CTR prediction.
        }
        \label{fig:ctr_model_with_cross_gnn}
          \vspace{-2em}
\end{figure}

\subsubsection{CrossNet}
% With the help of node encoding module, we can represent each node with an  neighbor aware embedding  $h_{i}^{(K)}$.
The CrossNet transforms the neighbor aware node-level embedding $h_{i}^{(K)}$ into a edge (i.e., cross feature) space and gives a prediction for the edge attribute.
% Then to characterize the cross features for each edge, a Feature CrossNet is
% propose, where we feed the embeddings of the pair nodes in each edge to the feature cross module and give a prediction for the edge attribute.
Formally, for each edge $e_{u,v}$, the edge attribute can be predicted by,
\begin{equation}
p_{u,v} = \mathrm{CrossNet}(h_{u}^{(K)},h_{v}^{(K)})
\end{equation} 
where $p_{u,v}$ is the predicted edge attribute, and $\mathrm{CrossNet}$ can be a simple dot production or a multilayer perceptron (MLP). 
In this paper, we implement $\mathrm{CrossNet}$ as a single perceptron layer.

\subsubsection{Weighted Square Loss}
\label{sec:Weighted Square Loss}
When obtaining the edge attribute prediction $p_{u,v}$, we can simply apply a square loss to train the PCF-GNN, i.e., 
% \begin{equation}
$loss = \sum_{e_{u,v}}||p_{u,v}-a_{u,v}||^{2}$
    % \label{eqn:loss}
% \end{equation}
. 
% By minimizing this loss, $p_{u,v}$ can directly capture the cross features explicitly. 
The problem is that, in this loss, each edge takes as an  equal contribution to the loss optimization.
However, different edges take different importance in CTR tasks.
For example, the pair \textit{<basketball player, "Nike-Air Jordan">} may appear more times than the pair \textit{<basketball player, mouse>} in history.
% We want to 
If we treat these two pairs equally in loss calculation, the learned PCF-GNN may be difficult to distinguish which kind of cross features is important for the interaction modeling and may mislead the CTR model to recommend a mouse to a basketball player.
To address this problem, we propose a simple but effective weighted square loss.
The key idea is allocating different weight to different edges by their co-occurrence,
% and we can rewrite Equ \ref{eqn:loss} as,
\begin{equation}
    loss = \textstyle\sum_{e_{u,v}}log(Count(u,v)+t)||p_{u,v}-a_{u,v}||^{2}
    \label{eqn:weight_loss}
\end{equation} where $log$ is a smooth operation which is used to avoid an extremely large $Count(u,v)$, and $t$ is a small positive constant ($t=1$ as default) to avoid a zero weight produced by $log$.
% Note when $C(u,v)=1$, the corresponding edge can be taken as a

% The pipeline of online service are shown in Fig.\ref{fig:online_service}. 

% Note that pre-trained CrossGNN are easy to unite with downstream tasks.

% We show the pseudo-code of CrossGNN in Algorithm \ref{alg:GSKN}. 

\subsection{PCF-GNN Application}
\label{sec:PCF-GNN Application}
When PCF-GNN is pre-trained, the learned knowledge about the explicit semantic information can be well characterized by $p_{u,v}$.
Then we can use this knowledge to help the downstream task, especially for the CTR prediction task.
Here we take the CTR prediction task as an example to show how to combine PCF-GNN with downstream models.
% In general, the CTR prediction model can be taken as a Eembedding\&MLP based deep model \cite{cheng2016wide,guo2017deepfm,deepcrosssing} where the embedding vectors of different features are concatenated together and then this concatenated vector is feed into a MLP to predict the click label.
% Then we can take the predicted explicit cross feature $p_{u,v}$ of PCF-GNN as an additional feature vector, and concatenate it with embedding vectors of features in CTR prediction model.
% Specifically, 
As shown in Fig \ref{fig:ctr_model_with_cross_gnn}, the right part is a standard CTR model which follows an Embedding\&MLP architecture \cite{cheng2016wide,guo2017deepfm,deepcrosssing}.
% where the embedding vectors of different features are concatenated together and then this concatenated vector is feed into a MLP to predict the click label.
Then we can take the predicted $p_{u,v}$ as an additional feature, and concatenate it with the embeddings of the CTR model (as shown in the dotted part in Fig \ref{fig:ctr_model_with_cross_gnn}).
In this stage, by default, we drop the node encoding module for the efficient purpose and directly keep the output $h_i^{(K)}$ of this module as the pre-trained embedding and the parameters in PCF-GNN are fixed.
(Note when considering the \emph{fine-tuning} strategy, the complete PCF-GNN including the node encoding module needs to be introduced in the CTR model).
In this way, the explicit semantic cross features learned by PCF-GNN can help to improve the performance of the CTR model.

\subsection{Discussion}
\label{sec:Theoretical Analysis}

In this section, we provide a brief discussion of our model on addressing the two challenges mentioned in Section \ref{sec:Introduction}.
% its efficient and effective ability of modeling explicit cross features. 

\noindent\textbf{Generalization.}
In PCF-GNN, the explicit semantic cross features (i.e., edge attribute) are inferred by pre-trained node embeddings.
It can infer the explicit semantic cross features of new pairs (i.e., new edges).
For example, even if there have no interactions between a programmer and the latest "Nike-Air Jordan" in history, PCF-GNN can still infer the new cross pair \textit{<programmer, the latest "Nike-Air Jordan">} by the pre-trained node embeddings of the programmer and the latest "Nike-Air Jordan", which finally can help the recommendation of the latest "Nike-Air Jordan" to a programmer user. 

\noindent\textbf{Low memory cost.}
As introduced in Section \ref{sec:Introduction}, theoretically, the memory cost of statistical explicit semantic cross features is $O(N_1N_2)$.
While the main cost of inferred explicit cross features comes from the pre-trained node embedding table, i.e., $O((N_1+N_2)d)$ where $d$ is the embedding dimension.
% and $(d+1)$ refers to store the key (e.g., node ID) and the embedding vector.
% Let $N_s$ and $|V|=N_u + N_i$ denote the number of rows for pairwise statistical features, number of users and items which participant in graph construction. 
% The dimension of are 
Considering $N_1$ and $N_2$ are usually a large number in the web-scale application, we can reduce the memory cost from $O(N_1N_2)$ to $O((N_1+N_2)d)$.

% \noindent\textbf{Time Complexity}

% \noindent\textbf{Relationship with SVD}
% Singular value decomposition (SVD) \cite{stewart1993early} is a collaborative filtering method for recommendation. 
% It aims at providing users with items’ recommendation from the latent features of item-user matrices. 
% Here we show some theoretical relationships between our model and SVD.
% We first take a brief look at SVD \cite{stewart1993early},
% $$M=U\Sigma V$$
% where $M\in R^{m\times n}$ is the ctr of user-item matrix, with $m$ users and $n$ items.
% $$M_{i,j}=f(U_{i,k})\Sigma f(V_{k,k})$$
% where $f(\cdot)$ is the non-linear 2-layer GNN networks.

\section{Experiments}
% In this section, we move forward to evaluate the effectiveness of our proposed approach. We aim to answer the following questions:
% \begin{itemize}
%     \item \textbf{(Q1)} Whether the graph construction process is effective and lightweight ?
%     \item \textbf{(Q2)} How does our proposed PCF-GNN perform in capturing both explicit and implicit feature interactions?
%     \item \textbf{(Q3)} What are the pre-trained PCF-GNN effects on various downstream tasks ?
%     \item \textbf{(Q4)} How much storage space does PCF-GNN need in online servering service ?
%     \item \textbf{(Q5)} How does the settings of networks influence the performance of PCF-GNN ?
% \end{itemize}

% \begin{table}[t]
%     \centering
%     \begin{tabular}{cccccc}
%     \toprule
%     Dataset & {Type} & 
%     {User} &
%     {Item} & Samples & Fields\\
%     \hline
%     Alimama & Industrial & 0.36 B & 0.15 B  &  20 B & 100+ \\
%     % Avazu & Public & 2,686,408  & 4,737 & 40,428,967 & 23\\
%     MovieLens & Public & 6040 & 3706 & 1000209 & 10 \\
%     \bottomrule
%     \end{tabular}
%     \caption{Dataset statistics and properties. B refers to Billion.}
%     \label{tab:datasets}
% \end{table}

\subsection{Experiment Setup}
% We first introduce the datasets, comparison methods as well as settings for experimental evaluation. 
\noindent \textbf{Datasets.} We conduct experiments on a large-scale industrial \textbf{Alibaba} dataset
% \footnote{http://taobao.com/} 
(containing 60 billion samples) collected by the online platform and a public \textbf{MovieLens} dataset
% \footnote{https://grouplens.org/datasets/movielens/} 
(containing 1000209 samples). 
Note each dataset is split into pre-trained data used for constructing graphs, downstream training data, and downstream test data respectively.
% Detailed statistics and properties are listed in Table \ref{tab:datasets}. 
% \begin{itemize}
%     \item \textbf{Alibaba\footnote{http://taobao.com/} dataset}. The co-occurrence commodity graph is constructed using users’ click behavior data during 30 days. There are 1 billion nodes and 8 billion edges. Types of graph node are all commodities. And the average output degree of graph node is 4.
%   \item \textbf{MovieLens\footnote{https://grouplens.org/datasets/movielens/}} This dataset contains users’
% ratings on movies. During binarization process, we treat samples with a rating less than 3 as negative samples because a low score indicates that the user does not like the movie. And we treat samples with a rating greater than 3 as positive samples.
% \end{itemize}

\noindent \textbf{Baselines.} 
We compare PCF-GNN with the following baselines.
% \begin{itemize}
% \noindent
% $\bullet$ 
(1) Implicit semantic modeling methods, including Wide\&Deep 
\cite{cheng2016wide}, DeepFM \cite{guo2017deepfm}, AutoInt \cite{autoint}, FiBiNET \cite{huang2019fibinet}.
% These methods are used as the CTR models.
% \noindent
% $\bullet$ 
(2) GNN-based methods that capture implicit semantic cross features, including Fi-GNN \cite{fgcnn}.
% \noindent
% $\bullet$
(3) Graph pre-trained models, which are originally designed to capture neighbor structure rather than explicit semantic cross features, i.e., GraphSAGE \cite{hamilton2017inductive} and PGCN \cite{hu2019strategies}.
% Considering the constructed graphs are based on the interactions among features
% Although these methods are originally designed to capture graph structure rather than explicit cross features
% , we adjust these methods into our pre-training pipeline and adopt a link prediction task to allow them 
% To show the effectiveness of proposed pre-train task on  
% \end{itemize}
    % \item is a graph-level pre-trained gcn, using masked edge. It is designed for general graph structured data.
Note methods in (1)(2) can be directly taken as the downstream models for the CTR task.

\noindent \textbf{Hyperparameter Settings.} We take 8-dimensional embeddings for all pre-trained methods. Parameters of other baselines follow recommended settings in relevant codes. 
For PCF-GNN, we take 2-layer networks with a hidden layer sized 64.

\subsection{Performance on CTR Task}
\label{sec:Performance on CTR task (Q2)}
\subsubsection{Performance on the public dataset.}
To investigate whether PCF-GNN captures explicit semantic cross features and further improves the performance, we conduct the experiments on the downstream CTR task where the additional features generated by SESCF or pre-trained models are combined into the downstream models (similar to Fig \ref{fig:ctr_model_with_cross_gnn}).
% The CTR results are listed in Table \ref{tab:ctr}. 
Here we construct the graph based on the interactions between User\_id  and Genres on MovieLens.
There is a total of 60,574 edges and 5,992 nodes. 
Then we use AUC \cite{bradley1997use} as the metric (Note it is noticeable that a slightly higher AUC at \textbf{0.001-level} is regarded significant for the CTR task \cite{zhou2018deepdin,autoint}). 
% In addition, to further show the effectiveness of PCF-GNN, we select the samples from test data which not appeared in the statistical features. 
% Results are shown in Table \ref{tab:no_apperance_res}.
Results are reported in Table \ref{tab:ctr} (Note ESCF refers to explicit semantic cross features).
We summarize the following findings:
% \begin{itemize}
%     \item Our proposed PCF-GNN outperforms all comparison baselines in almost all CTR models, which demonstrates the effectiveness of our model.
%     \item Compared with No ECF, all modeling explicit cross features methods are able to improve CTR results in general, which shows the usefulness of explicit cross features.
%     \item Almost all pre-trained models gains better performance, against statistical cross features. It shows that introducing pre-trained pipeline is helpful in recommendation systems.
%     \item Compared with other GNN models, PCF-GNN gains the best performance. It illustrates that our proposed loss is able to capture implicit features interactions.
%     \item  Note that PCF-GNN gains 0.01 improvement on the FGCNN, thus shows that PCF-GNN is able to capture hidden and implicit features interactions. 
% \end{itemize}
% \begin{itemize}
    % \item     
% $\bullet$ 
(1) The proposed PCF-GNN outperforms all baselines, which shows the effectiveness of our model.
    % \item 
% $\bullet$ 
(2) Compared with No ESCF, even if most of the downstream models capture cross features implicitly, the models with ESCF (including SESCF and PCF-GNN) are able to improve CTR results, which shows the usefulness of explicit semantic modeling.
    % \item Almost all pre-trained models gains better performance, against SECF. It shows that introducing pre-trained pipeline is helpful in recommendation systems.
    % \item
% $\bullet$
(3) Compared with SESCF, PCF-GNN can further improve the CTR performance. 
It shows PCF-GNN has the superiority of explicit semantic modeling.
    % \item
% $\bullet$
(4) Compared with other pre-trained models (i.e., GraphSAGE and PGCN), although due to capture the structure of the constructed interaction-based graph they can improve CTR performance, PCF-GNN achieves the best performance. 
It illustrates the design of our pre-training task and our model can capture explicit semantic cross features better.
    % \item  Note that PCF-GNN gains 0.01 improvement on the FGCNN, thus shows that PCF-GNN is able to capture hidden and implicit features interactions. 
% \end{itemize}

\begin{table}[t]
\centering
\begin{adjustbox}{max width=\linewidth}
\begin{tabular}{cccccccc}
\hline
% \multirow{3}{*}{Task} & \multirow{3}{*}{\begin{tabular}[c]{@{}c@{}}Downstream  \\ Model\end{tabular}} & \multirow{3}{*}{No ECF} & \multicolumn{5}{c}{ECF}                                     \\ \cline{4-8} 
%                       &                        &                         & \multirow{2}{*}{SECF} & \multicolumn{4}{c}{Pre-trained model} \\ \cline{5-8}
\begin{tabular}[c]{@{}c@{}}Downstream  \\ Task\end{tabular} & \begin{tabular}[c]{@{}c@{}}Downstream  \\ Model\end{tabular} &No ESCF & SESCF & GraphSAGE &PGCN &PCF-GNN \\ \hline 
                      % &                        &                         &                     & LINE    & GCN     & PGCN    & PCF-GNN  \\ \hline
\multirow{6}{*}{CTR}  & Wide\&Deep             & 0.7144                  & 0.7253                & 0.7254  & 0.7259  & \textbf{0.7266}   \\ 
                      & DeepFM                 & 0.7238                  & 0.7268                & 0.7285 & 0.7288  & \textbf{0.7293}   \\ 
                      & FiBiNET                & 0.7171                  & 0.7210                & 0.7251  & 0.7254  & \textbf{0.7261}   \\  
                      & AutoInt                & 0.7238                  & 0.7280       & 0.7276  & 0.7280  & \textbf{0.7282}   \\ 
                      % & FNN                    & 0.7251                  & 0.7258              & \textbf{0.7289}  & 0.7270  & 0.7276  & 0.7280   \\ 
                    %   & FGCNN                  & 0.6975                  & 0.7019                & 0.7067  & 0.7075  & \textbf{0.7090}   \\ 
                      & Fi-GNN                  & 0.7256                 & 0.7301                & 0.7277 & 0.7283  & \textbf{0.7315}   \\ 
                      \hline \hline 
Task1                   & LightGBM                    & 0.7110                  & 0.7107                 & 0.7126  & 0.7127  & \textbf{0.7239}   \\ \hline 
Task2                   & LightGBM                    & 0.5791                  & 0.5862              & 0.5839  & 0.5931  & \textbf{0.5990}   \\ \hline
\end{tabular}
\end{adjustbox}
\caption{Results of different tasks on MoiveLens dataset. 
}
  \label{tab:ctr}
  \vspace{-2em}
\end{table}

\subsubsection{Performance on the in-house industrial dataset}
We further conduct experiments on an industrial dataset (Alibaba).
Here, we construct a heterogeneous graph based on 11 kinds of cross features.
There is a total of 20 billion edges and 0.7 billion nodes.
The results are reported in Table \ref{tab:online_res}.
Note DNN refers to Wide \& Deep \cite{cheng2016wide} which is taken as the CTR model.
We can find that PCF-GNN still achieves the best performance.
It indicates that PCF-GNN can well model the explicit semantic cross features.

% \begin{table}[htbp]
%     \centering
%     \begin{tabular}{cccc}
%     \toprule
%     Model &  {AUC} \\
%     \hline
%     DNN &  0.7442 \\
%     DNN + SECF & 0.7466 \\ 
%     DNN + PCF-GNN & \textbf{0.7477}\\
%     \bottomrule
%     \end{tabular}
%     \caption{CTR performance on industrial alibaba dataset.}
%     \label{tab:online_res}
% \end{table}

\begin{table}[t]
    \centering
    % \begin{adjustbox}{max width=\linewidth}
    \scalebox{0.9}{
    \begin{tabular}{cccc}
    \toprule
    Model &  DNN&DNN + SESCF&DNN + PCF-GNN \\
    \hline
    AUC&0.7442  & 0.7466& \textbf{0.7477}\\
    \bottomrule
    \end{tabular}
    }
    % \end{adjustbox}
    \caption{CTR performance on industrial Alibaba dataset.}
    \label{tab:online_res}
    \vspace{-3em}
\end{table}

\subsection{Performance on Other Downstream Tasks}
% Apart from CTR prediction task, 
To show the universality of PCF-GNN, we evaluate PCF-GNN on various related downstream tasks.
Detailedly, we select some features (i.e., age and occupation) on MovieLens as labels (Note other features can also be considered as labels) and conduct experiments on predicting the corresponding labels, i.e., predicting user's age (defined as Task1) and user's occupation (defined as Task2).
We use LightGBM \cite{ke2017lightgbm} as the downstream classifier, and accuracy as the metric.
% The age and occupation tasks have 3 and 7 classes respectively. 
% The pre-trained models will provide additional explicit cross features, compared with the original data.  
Results are shown in Table \ref{tab:ctr}. 
We observe that PCF-GNN significantly improves the results of all downstream tasks, compared with other competitors.
It shows the effectiveness of the learned explicit cross feature in various related downstream tasks.

\subsection{Evaluation of Model Generalization}
% As introduced in Section \ref{sec:Introduction}, statistical explicit cross feature lacks of generalization.
Here we conduct experiments to evaluate the generalization of models (addressing challenge one).
Specifically, we first calculate the Hit Rate (HR) of explicit semantic cross features on test data of MovieLens where there may exist some new cross feature pairs.
Then we evaluate the AUC of different methods on the test samples which only contain new cross feature pair (defined as New).
We also provide the results on the whole test samples (defined as Org) for comparison. 
The results are reported in Table \ref{tab:no_apperance_res}.
Note DNN refers to Wide \& Deep \cite{cheng2016wide} and $\Delta_{XX}$ means the gap with DNN on the test dataset $XX$.
We can find that 
(1) Since SESCF cannot infer the explicit semantic cross feature of new pairs, SESCF can cover only 78.27\% of test samples. 
While PCF-GNN can cover 91.81\% of test samples.
 % can be guaranteed by explicit cross feature due to the generalization ability.
(2) SESCF shows a negative impact on the New dataset.  
While PCF-GNN can always achieve improvement both on the New and Org datasets.
These demonstrate that PCF-GNN has a good generalization ability and can infer the explicit semantic cross feature of new pairs to improve the model performance.

\begin{table}[t]
    \centering
    \scalebox{0.8}{
    \begin{tabular}{cccccc}
    \toprule 
    % \multirow{2}{*}{Model} &
    % \multicolumn{2}{c}{==}& 
    % \multicolumn{2}{c}{$\Delta$} \\
    % \cmidrule(lr){2-3}
    % \cmidrule(lr){4-5}
    Model & HR & New &$\Delta_{New}$ & Org &$\Delta_{Org}$\\
    \hline
    DNN & --&0.6972 & -- & 0.7144 & -- \\
    DNN + SESCF &78.27\% & 0.6916 & -0.0056 & 0.7253 & 0.0109 \\ 
    DNN + PCF-GNN & \textbf{91.81}\% & \textbf{0.7010} & \textbf{0.0038} & \textbf{0.7266} & \textbf{0.0122} \\
    \bottomrule
    \end{tabular}
    }
    \caption{Results about generalization. }
    \label{tab:no_apperance_res}
    \vspace{-2em}
\end{table}

\subsection{Evaluation of Memory Cost}
% Storage revenue of online server has always been a crucial step in online recommender systems. 
In this section, we conduct experiments on the industrial Alibaba dataset to analyze the memory cost of PCF-GNN (addressing challenge two). 
% For simplicity, since the elements in the <k,v> table of SECF and the pre-trained embedding table of PCF-GNN is proportional to the memory cost, we count the number of these elements for memory comparison.
The memory cost of modeling the explicit semantic cross features of different methods is reported in Table \ref{tab:online_storage_2}.
With the help of PCF-GNN, we can save 56\% memory cost compared with SESCF which efficiently improves memory storage.
Note when each feature has interactions with all other features, theoretically, the memory cost can be reduced from $O(N_1N_2)$ to $O((N_1+N_2)d)$.

\begin{table}[t]
    \centering
    \scalebox{0.8}{
    \begin{tabular}{cccc}
    \toprule
     Method & SESCF & PCF-GNN  & Change\\
    \hline
    Memory Cost (GB) & 84& 37 & -56\% \\
    \bottomrule
    \end{tabular}
    }
    \caption{Memory cost of different methods.}
    \label{tab:online_storage_2}
    \vspace{-2em}
\end{table}

% \begin{table}[htbp]
%     \centering
%     \begin{tabular}{cccc}
%     \toprule
%      &  & 
%     AUC & LogLoss \\
%     \hline
%     \multirow{2}{*}{Loss} & weighted & \textbf{0.7266}   & \textbf{0.4637}\\
%     & unweighted & 0.7257 & 0.4683 \\
%     \hline
%     \multirow{2}{*}{Fine-tune} & with & \textbf{0.7268} & \textbf{0.4563}\\
%     & without & 0.7266 & 0.4637\\
%     \bottomrule
%     \end{tabular}
%     \caption{Results with different model components.}
%     \label{tab:para4}
% \end{table}

\subsection{Ablation Study}
We conduct experiments to analyze the influence of Weighted Loss (Section \ref{sec:Weighted Square Loss}), Fine-Tune (Section \ref{sec:PCF-GNN Application}), and the GNN architecture in the node encoding module (Section \ref{sec:Node Encoding Module}), to provide better understandings of PCF-GNN.
Specifically, we conduct experiments on MovieLens with Wide \& Deep as the CTR model to verify the analysis.
Results are shown in Table \ref{tab:para4} (Note WL refers to weighted loss and FT refers to fine-tune. Base refers to PCF- GNN without GNN, WL, and FT). 
We conclude that the weighted loss, Fine-tune, and GNN can improve the model performance.
% have better performances than unweighted loss and without fine-tune scheme respectively. 

% Then we investigate the performance with the change of model depth and embedding size $d$ in Fig. \ref{fig:paras} (a) and (b). 
% The performance about model depth continuously increase when number of layers reaches two. Then the performance becomes stable, showing that adding extremely high-order features are not informative. In addition, larger $d$ will improve the performance of PCF-GNN.

% \begin{figure}[htbp]
% \centering
%  \subfigure[Hidden Size]{

%  \includegraphics[width=0.48\linewidth]{imgs/hidden_size.pdf}}
%          \subfigure[Depth of the layer]{ \includegraphics[width=0.48\linewidth]{imgs/layers.pdf}}
%         \caption{Parameter analysis on PCF-GNN.}
%         \label{fig:example}
% \end{figure}

\begin{table}[t]
\scalebox{0.9}{
    \centering
    \begin{tabular}{ccccc}
    \toprule
     &  Base & Base+GNN & Base+GNN+WL & Base+GNN+WL+FT\\
    \hline
    AUC&0.7256&0.7261&0.7266&0.7268\\
    \bottomrule
    \end{tabular}
}
    \caption{Results of the ablation study.
    % Note WL refers to weighted loss and FT refers to fine-tune.
    % Base refers to PCF-GNN without GNN, WL and FT.
    }
    % For simplicity, WL refers to weighted loss and FT refers to fine-tune.}
    \label{tab:para4}
    \vspace{-3em}
\end{table}

\section{Conclusion}
In this paper, we propose PCF-GNN, a pre-trained recommendation model based on GNNs. 
We first transform the interactions among features in history into an informative graph.
Then we introduce pre-trained GNN models to learn efficient and effective explicit relationships among features. 
We propose a novel training loss to capture explicit semantic cross features. 
Experiments on both online and public datasets show our model is competent in both performance and memory-efficiency.
\newpage

% \bibliographystyle{ACM-Reference-Format}

\end{document}